\documentclass[runningheads]{llncs}
\usepackage[T1]{fontenc}
\usepackage{graphicx}
\usepackage[utf8]{inputenc}
\usepackage[nocompress]{cite}
\usepackage{graphbox}
\usepackage[nice]{nicefrac}
\usepackage[cmex10]{amsmath}
\usepackage{amssymb}
\usepackage{amsfonts}
\usepackage{bbding}
\usepackage{algorithmic}
\usepackage{url}
\usepackage{textcomp}
\usepackage{comment}
\usepackage{gensymb}

\usepackage{siunitx}
\usepackage{paralist}
\usepackage{tikz}
\usetikzlibrary{patterns}
\usepackage{tkz-euclide}
\usetikzlibrary{arrows}
\usetikzlibrary{positioning, shapes.geometric, fit}

\usepackage{enumitem}
\usepackage{rotating}
\usepackage{afterpage}
\usepackage{tablefootnote}
\usepackage[shortcuts]{extdash}
\usepackage[margin=1in]{geometry}
\usepackage{xcolor}
\usepackage{hyperref}
\newcommand{\QuotationMarks}[1]{``{#1}''}

\newcommand{\secref}[1]{\hyperref[#1]{Section~\ref*{#1}: \nameref*{#1}}}
\newcommand{\secrefplain}[1]{\hyperref[#1]{\ref*{#1} \nameref*{#1}}}
\newcommand{\secrefnumber}[1]{Section~\hyperref[#1]{\ref*{#1}}}
\newcommand{\secrefsnumber}[2]{Sections~\hyperref[#1]{\ref*{#1}} and~\hyperref[#2]{\ref*{#2}}}

\newcommand{\myvector}[1]{{\boldsymbol{\mathbf{#1}}}}

\begin{document}
	
\title{Rail Track Extraction from\\Rasterized Classified Point Clouds Using a Full-Resolution,\\Fully Convolutional Recurrent Neural Network}
\titlerunning{Rail Track Extraction from Classified Point Clouds}
\author
{
	Alexander Gribov\Envelope
	\and 
	Jie Chang
}
\authorrunning{A. Gribov and J. Chang}
\institute{Esri, 380 New York Street, Redlands, CA 92373-8100, USA \\
	\email{agribov@esri.com; jchang@esri.com}}

\maketitle

\begin{abstract}
	Rail track extraction is essential for effective railway asset management and maintenance, especially in automated inspection and mapping workflows. This paper introduces a novel method for extracting rail tracks from classified 3D point clouds using a fully convolutional recurrent neural network that preserves full spatial resolution and is trained exclusively on synthetically generated data. This approach enhances per-pixel quality and is particularly suited for rail track extraction. The proposed method begins by rasterizing points corresponding to railroad tracks, then applies the neural network to reduce noise and yield a cleaner track representation suitable for vectorization \cite{VectorizationAndParityErrors}. Subsequent morphological operations further refine the resultant data, enabling accurate track centerline extraction. Next, the extracted centerlines undergo smoothing to eliminate residual irregularities \cite{Smoothing, SmoothingNetwork}. Finally, the algorithm transfers 3D information from lidar points onto 2D polylines and applies additional vertical smoothing. A single centerline for both tracks is found using the Dynamic Time Warping (DTW) algorithm \cite{DynamicTimeWarpingAlgorithmReview}. The final outcome consists of rail top centerlines and track centerlines derived for rail pairs, with minimal manual intervention. Experimental validation confirms the effectiveness of this method in yielding high-quality rail track extraction.
\end{abstract}

\section{Introduction}

Accurate rail track information is critical for safe and efficient rail operations~\cite{AutomatedExtractionof3DRailwayTracksfromMobileLaserScanningPointClouds}. In particular, precise track centerlines\textemdash located midway between the two rails\textemdash are essential for systems such as Positive Train Control (PTC), which monitor train location and speed to automatically prevent collisions, overspeed derailments, and unauthorized movements through misaligned switches~\cite{PositiveTrainControl}.

To meet the precision and safety requirements of modern rail systems, detailed 3D information about track geometry must be acquired and regularly updated. Mobile lidar scanners mounted on rail vehicles are capable of capturing dense, high-resolution point clouds at high acquisition rates, making them well-suited for mapping rail corridors, including tracks, switches, crossings, and other critical infrastructure components~\cite{ThreeEmergingLIDARApplicationsForRailDataCapture}.

However, extracting rail points from large-scale, high-density lidar datasets remains a significant challenge, especially in complex or cluttered environments~\cite{AFastAlgorithmForRailExtractionUsingMobileLaserScanningData}. Manual inspection and semi-automated workflows can be time-consuming, and rule-based extraction methods often require careful tuning to specific sensors, rail layouts, and scene conditions. In contrast, deep learning offers a powerful alternative by learning patterns directly from data, enabling more automated and potentially more accurate rail classification. Recent studies~\cite{SemanticSegmentationOfPointCloudsWithPointNetAndKPConvArchitecturesAppliedToRailwayTunnels, PointCloudSemanticSegmentationOfComplexRailwayEnvironmentsUsingDeepLearning, MultimodalDeepLearningForPointCloudPanopticSegmentationOfRailwayEnvironments} have demonstrated promising results in applying deep learning to classify rail environments from point clouds. Yet, most existing methods focus on broader scene classification, while only a few are dedicated specifically to extracting rail features themselves.

Even after rail points are classified, modeling them as 3D polylines introduces additional challenges. This step requires vectorization and smoothing, often relying on complex algorithms to address noise, gaps, curvature variations, and rail direction at intersections~\cite{AutomaticExtractionOfRailroadCenterlinesFromMobileLaserScanningData, RailTrackDetectionAndProjectionBased3DModelingFromUAVPointCloud, FullyAutomatedMethodologyForTheDelineationOfRailwayLanesAndTheGenerationOfIFCAlignmentModelsUsing3DPointCloudData, FullyAutomatedExtractionOfRailtopCenterlineFromMobileLaserScanningData}. These processes are computationally intensive and sensitive to data quality, making them difficult to scale across diverse rail systems.

To address these challenges, we explore a synthetic-data-trained deep learning approach for rail vectorization from rasterized classified point clouds. We propose a novel method, Full-Resolution Progressive Dilated Fusion (FRPDF), for extracting 3D rail polylines and centerlines from classified rail points. To our knowledge, this specific combination of synthetic raster training, full-resolution recurrent dilated fusion, and rail-vectorization postprocessing has not been previously applied to rail track extraction. Our approach leverages a neural network trained on synthetic raster data, which eliminates dependence on real-world rail datasets\textemdash often proprietary and restricted by rail companies. Using synthetic data enables the generation of diverse training samples representing varying rail widths, curves, gaps, and intersection configurations. To simulate real-world conditions, we introduce multiple levels of noise into the synthetic rasters. This significantly reduces the need for manual labeling while improving the model's generalizability and robustness.

In addition, working with raster data simplifies the input structure compared to raw point cloud data and lowers computational costs. This allows us to build our model on top of established neural network architectures designed for raster inputs, offering greater flexibility and development efficiency.

The proposed FRPDF network operates at full resolution to preserve fine geometric details and employs a reduced number of channels to minimize computational overhead. In contrast to downsampling architectures styled after U-Net, FRPDF preserves full spatial resolution, which is advantageous for rail vectorization. FRPDF utilizes a lightweight encoder-decoder design with fused channel strategies that is well suited to handling high-noise scenarios. We conducted an exploratory comparison with Asymmetric Convolutional Neural Network (ACNN)~\cite{ACNN} during development. However, because a fully fair parameter- and capacity-matched comparison would require substantial reengineering, we do not present this comparison as a definitive benchmark in this paper.

In summary, the main contributions of this paper are the following:
\begin{enumerate}
	\item FRPDF, a novel deep learning architecture for rail vectorization from classified point clouds
	\item A synthetic raster-based training strategy that avoids reliance on proprietary rail data and reduces the need for manual labeling
	\item Experimental evaluation of FRPDF on synthetic test rasters, together with qualitative validation on real lidar data
\end{enumerate}
The proposed architecture was developed from a design process inspired by U-Net, but the final full-resolution formulation has conceptual similarities to ACNN while avoiding the limitations of downsampling-based architectures.

All preprocessing, labeling, training, classification, and extraction of rails was performed in ArcGIS Pro~3.7. 

Although this work focuses on rail track extraction, the proposed FRPDF solution may also be relevant to other rasterized reconstruction problems involving smooth, continuous rail-like structures observed with noise or gaps. The method is most appropriate when the target features follow strong geometric regularity, such as continuity, limited curvature variation, and predictable local shape. It is not intended as a general solution for arbitrary line extraction or, particularly, for structures with abrupt or random direction changes, where different architectures or task-specific models may be required.

The remainder of this paper is organized as follows. \secrefnumber{sec:RelatedWork} reviews related work. \secrefnumber{sec:DataAndRailClassification} describes the point-cloud dataset, preprocessing, and rail classification. \secrefnumber{sec:SyntheticDataGeneration} details the synthetic data generation process. \secrefnumber{sec:NeuralNetworkArchitecture} introduces the FRPDF architecture. \secrefnumber{sec:QualityEvaluation} presents the quality evaluation. \secrefnumber{sec:ApplyingTheNeuralNetworkToRealData} describes the application of the trained network to real lidar data. \secrefsnumber{sec:Vectorization}{sec:RailroadTrackCenterlineReconstruction} describe vectorization and railroad centerline reconstruction, respectively. \secrefnumber{sec:Example} presents an example, and \secrefnumber{sec:ConclusionAndFutureWork} concludes the paper and discusses future work.

\section{Related Work\label{sec:RelatedWork}}

Several methods have been proposed for extracting track centerlines or rail top centerlines from point cloud data.

Data-driven and model-driven approaches to extract track centerlines located between rails from mobile laser scanning (MLS) point clouds were introduced in~\cite{AutomaticExtractionOfRailroadCenterlinesFromMobileLaserScanningData}. Rail points were identified based on their height relative to the surrounding ground and the linear, parallel characteristics of rail geometry. In the data-driven approach, for each point on one rail, the algorithm searches for the nearby points on the opposite rail that are in a similar direction, fits a line through those points, and draws a perpendicular line from the original point to the fitted line. The midpoint is stored as a center point. These midpoints are then interpolated and smoothed into a continuous centerline using Random Sample Consensus (RANSAC). In the model-driven approach, a simplified parametric model, consisting of two rail pieces, is fitted to each rail segment using a Markov chain Monte Carlo (MCMC) algorithm. A Fourier series function is applied to interpolate a smooth and continuous rail model, and the final centerlines are derived as the geometric center between the modeled rails. Both approaches achieve high accuracy (\SIrange{2}{3}{\centi\metre}); the data-driven approach performs well near switches and small gaps, while the model-driven method is more robust for missing data and outliers. However, both approaches show reduced accuracy on curves.

A fully automated method for extracting track centerlines from MLS point clouds was proposed in~\cite{FullyAutomatedMethodologyForTheDelineationOfRailwayLanesAndTheGenerationOfIFCAlignmentModelsUsing3DPointCloudData}. The approach begins by associating point cloud data with the vehicle trajectory and filtering out points located far from the railway corridor. Additional filters based on height and surface roughness exclude elevated structures and irregular surfaces, significantly reducing noise from objects like trees, poles, and overhead cables. The remaining data is projected into 2D rasters representing height differences and intensity, allowing efficient isolation of potential rail regions. Initial rail tip points are roughly identified and refined using cylindrical templates to ensure alignment with the actual railhead and to eliminate interference from ballast and sleepers. Straight segments of track are detected using principal component analysis (PCA), and in each segment, transverse slices are taken at regular intervals to extract innermost railhead points. Midpoints between these pairs are computed and connected to form a smooth, continuous centerline polyline. The method supports complex rail layouts\textemdash including multiple tracks, crossings, and switches\textemdash while maintaining an average rail delineation error of less than \SI{3}{\centi\metre}.

A method for extracting railtop centerlines from MLS point clouds was proposed in~\cite{FullyAutomatedExtractionOfRailtopCenterlineFromMobileLaserScanningData}. The fully automated workflow first discards a majority of non-rail points through voxel-based pruning and subsequently identifies railhead points using local linearity and point-density filtering. Rail pairs are then detected within spatial voxels using a RANSAC-based model and aggregated into continuous track sections. Based on the detected rail pairs, approximate track centerlines are constructed, split into continuous paths, and subsequently smoothed using polynomial fitting. The smoothed track centerlines serve as reference paths for fitting railhead templates to the original point cloud, yielding the final railtop centerlines. The method achieves sub-centimeter accuracy on non-intersecting rails and comparable accuracy to existing methods in railway intersections while using only geometric information.

A method for detecting rail points and generating 3D rail models from photogrammetric point clouds generated from unpiloted aerial vehicle (UAV) imagery was proposed in~\cite{RailTrackDetectionAndProjectionBased3DModelingFromUAVPointCloud}; these point clouds are typically noisier and less dense than laser scanning data. Candidate rail points are first identified by detecting local height jumps within a grid. Nonrail points are then filtered out based on color intensity. Surface planarity is assessed using PCA, allowing the removal of irregular or nonplanar structures. A second-degree polynomial is then fitted to the candidate rail points using the M-estimator Sample Consensus (MSAC) algorithm, enabling the construction of smooth and continuous rail paths. To reconstruct the 3D rail model, the point cloud is projected onto the XZ plane and a predefined 2D model of the rail piece is fitted to the points. After fine-tuning the model's rotation and translation parameters, it is reprojected into 3D space. Finally, Fourier interpolation is applied to smooth transitions between segments, resulting in a continuous 3D rail model. While this method effectively reconstructs the rail surface from noisy UAV data, it does not address the extraction of accurate rail top centerlines from the refined model.

Overall, existing methods for rail geometry extraction differ significantly in data sources, algorithmic frameworks, and output objectives. Some approaches utilize MLS data, which offers high point density and precision, while others rely on UAV photogrammetric data, which is more accessible but typically noisier and less dense. Techniques span from data driven to model driven, with some targeting the extraction of track centerlines between rails while others focus on precise rail top centerlines. A few methods reconstruct rail models without explicitly generating centerlines. Across these approaches, a variety of fitting strategies\textemdash such as RANSAC, MSAC, polynomial regression, and template matching\textemdash are employed, each with its own advantages and limitations depending on data quality and scene complexity. Most workflows involve considerable parameter tuning and are sensitive to specific data characteristics. Given this complexity and variability, deep-learning-based approaches provide growing potential for technology to learn spatial patterns and automatically generate accurate centerlines with reduced manual effort and improved adaptability across diverse datasets.

\section{Data and Rail Classification\label{sec:DataAndRailClassification}}

\subsection{Dataset Description}
\begin{figure}[!htbp]
	\centering
	\includegraphics[width=\linewidth]{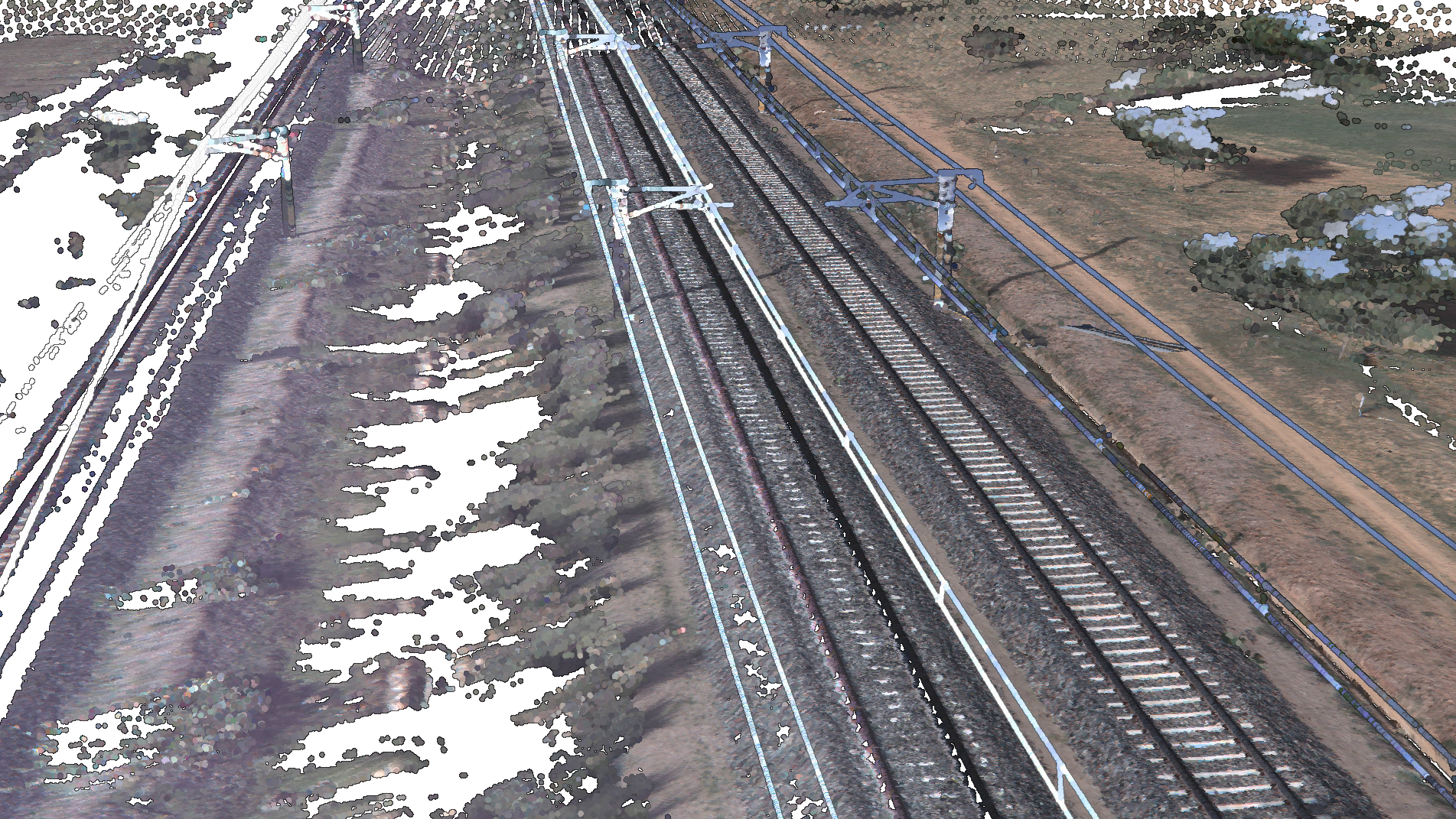}
	\caption
	{
		Point clouds of a multitrack railway corridor acquired using a train-mounted MLS system. Data courtesy of Esri India.
	}
	\label{fig:example_rail_tracks}
\end{figure}

Dense point clouds were acquired from a train-mounted MLS system along a multitrack railway corridor in India, as seen in Figure~\ref{fig:example_rail_tracks}. Due to the scanner's position on the rightmost track, this track and the adjacent center track exhibit high point density, capturing fine-scale rail geometry. In contrast, the leftmost track is sampled much more sparsely due to greater sensor-to-target range and oblique incidence angles, resulting in incomplete point coverage and reduced geometric fidelity.

The scene contains typical railway corridor features captured in MLS data, including rails, sleepers, ballast, and ground surfaces, along with vegetation growing between the parallel tracks. Overhead railway electrification infrastructure is also present, consisting of masts and supporting wires along the corridor. Adjacent terrain and roadside objects are partially captured. This dataset provides a representative MLS scenario for evaluating rail classification and extraction under different data quality conditions.

The track gauge is \SI{1.0}{\metre}, and each rail exhibits a thickness of \SI{61.91}{\milli\metre}. The railhead surface is typically \SI{0.15}{\metre} above the surrounding ground.

\subsection{Preprocessing and Rail Point Labeling}
All preprocessing and labeling was performed using interactive rail selection~\cite{ArcGISBlogRails} and class code editing tools in ArcGIS Pro. 

Processing boundaries were first delineated along the tracks, extending \SI{1.5}{\metre} outward on both sides to restrict the analysis to the immediate vicinity of the tracks. Ground points were classified using the Classify LAS Ground tool with the conservative option so that sleepers and ballast were also included as ground. The Classify LAS By Height tool was then used to identify low (< \SI{0.1}{\metre}) and high (> \SI{0.25}{\metre}) points relative to the ground; these were excluded from further processing as they fall outside the expected rail height zone.

Rail top points (points on the railhead) were labeled using the Interactive Rail Selection tool. Beginning with a small set of manually selected seed points, the tool iteratively expanded the selection along the track by evaluating neighboring candidates within a defined search radius. Points were added only if they satisfied geometric constraints characteristic of the rail: They had to fall within the expected rail width envelope, maintain minimum vertical separation from the ground, and remain within a narrow elevation tolerance relative to the prevailing railhead height. 

Finally, the Thin LAS tool was applied to thin ground points using a \SI{0.1}{\metre} grid and to remove low and high points outside the rail height zone. This significantly reduced the dataset size and improved the efficiency of subsequent model training. After preprocessing, the dataset consisted primarily of rail top points, thinned ground points, and unclassified rail-side points and intertrack vegetation within the \SIrange{0.1}{0.25}{\metre} height range.

\subsection{Model Training and Inferencing}

ArcGIS Pro was used for preparing training data~\cite{ReferencePointCloudToolsetPrepare}, model training~\cite{ReferencePointCloudToolsetTrain}, and inferencing~\cite{ReferencePointCloudToolsetClassify}. The labeled dataset was split into training and validation subsets. Training and validation samples were exported using the Prepare Point Cloud Training Data tool, which partitioned the data into \SI{4}{\metre} $\times$ \SI{4}{\metre} blocks (stored in H5 format) with a maximum of $3,000$ points per block to produce fixed-size inputs suitable for deep neural network training. When a block exceeded the point limit, the points were randomly partitioned into several blocks of point-limit size, with each point being used in no more than two blocks. The number of blocks was no more than necessary to ensure that each point appears in at least one block. For example, if a~block has $10,000$ points, four blocks of $3,000$ points will be generated.

Model training was conducted using the Train Point Cloud Classification Model tool with the RandLA\nobreakdash-Net architecture. RandLA\nobreakdash-Net is a lightweight and computationally efficient network designed for large-scale point cloud segmentation. It combines random sampling with local feature aggregation to effectively capture fine-scale geometric and texture information while remaining computationally efficient for large-scale point clouds.

Both training and validation loss exhibited a decreasing trend across epochs, indicating effective model convergence. The model achieved a validation recall of approximately $0.99$ for the rail class. 

Inference was performed using the Classify Point Cloud Using Trained Model tool. To maintain consistency across datasets, the same preprocessing and thinning procedures were applied to the test dataset prior to classification. The trained model successfully classified the majority of rail top points across the test area. However, several misclassification patterns were observed. A few low vegetation points were incorrectly classified as rail due to similarities in elevation and local geometry. Some rail top points on the leftmost track were missed, and classification ambiguity occurred along the upper rail edges where class boundaries were less distinct. Classification performance varied across tracks, reflecting the underlying data quality: The right and center tracks produced the most accurate results, while the left track showed reduced performance due to lower point density and increased geometric uncertainty.

\section{Synthetic Data Generation\label{sec:SyntheticDataGeneration}}

Synthetic data was generated through a systematic pipeline designed to simulate realistic railroad track conditions for training neural networks in rail extraction tasks. The process commenced with the generation of idealized railroad centerlines, followed by deliberate distortions and transformations to replicate various real-world imperfections and artifacts.

\subsection{Parameter Choices and Validation}
Track gauge and rail thickness parameters were defined by U.S. and European railway specifications. Synthetic rail thicknesses were varied randomly to simulate real-world variations encountered in practice. Rasterization cell size was selected to yield a rail representation spanning a few pixels, balancing resolution and computational efficiency.

\begin{figure}[!htbp]
	\centering
	\begin{tabular}{ccccc}
		\fbox{\includegraphics[width=0.18\linewidth]{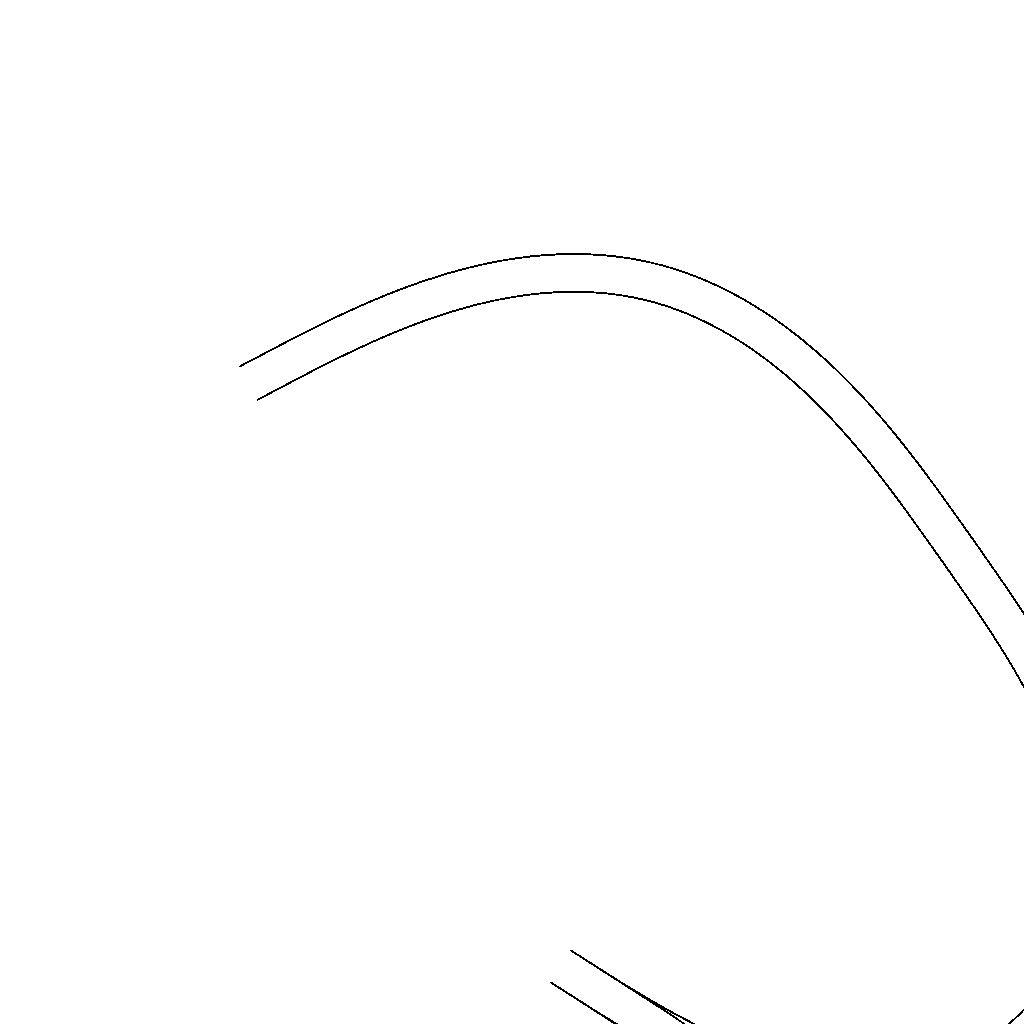}} &
		\fbox{\includegraphics[width=0.18\linewidth]{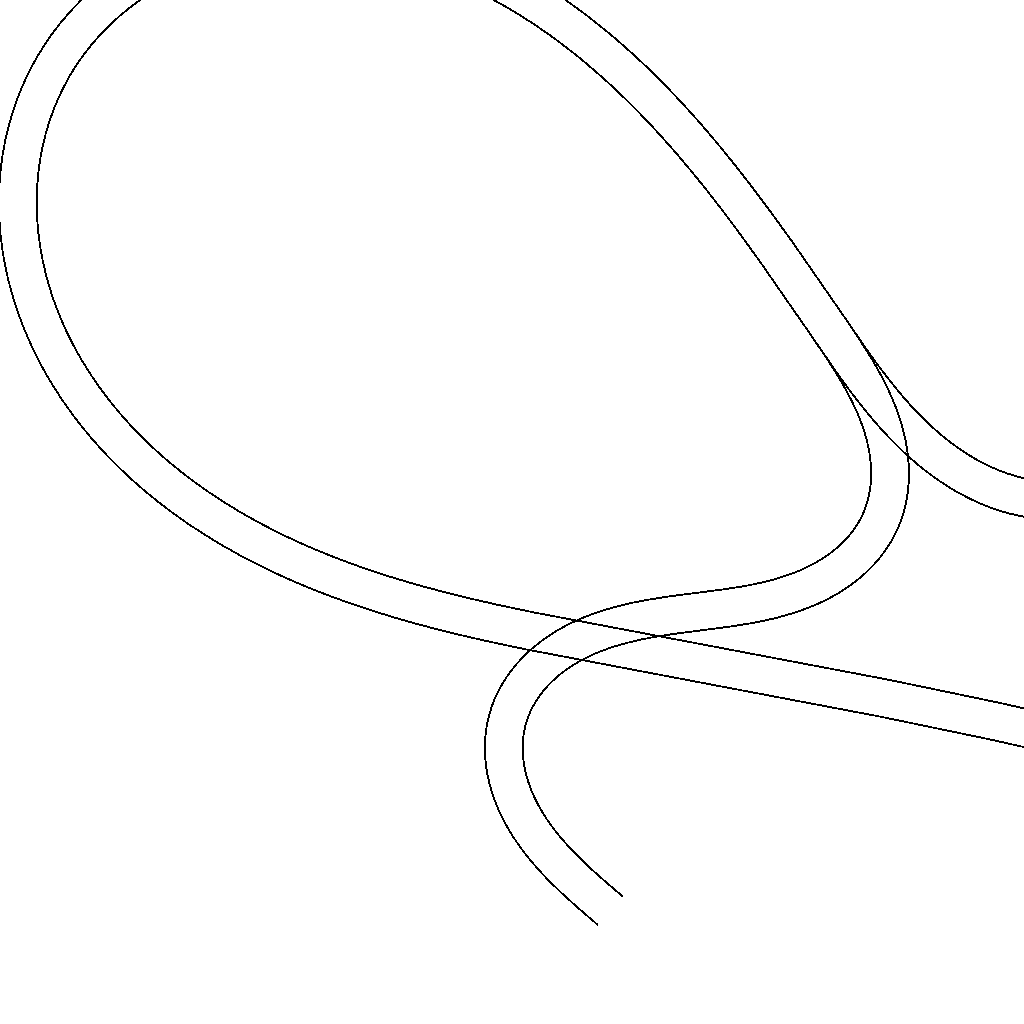}} &
		\fbox{\includegraphics[width=0.18\linewidth]{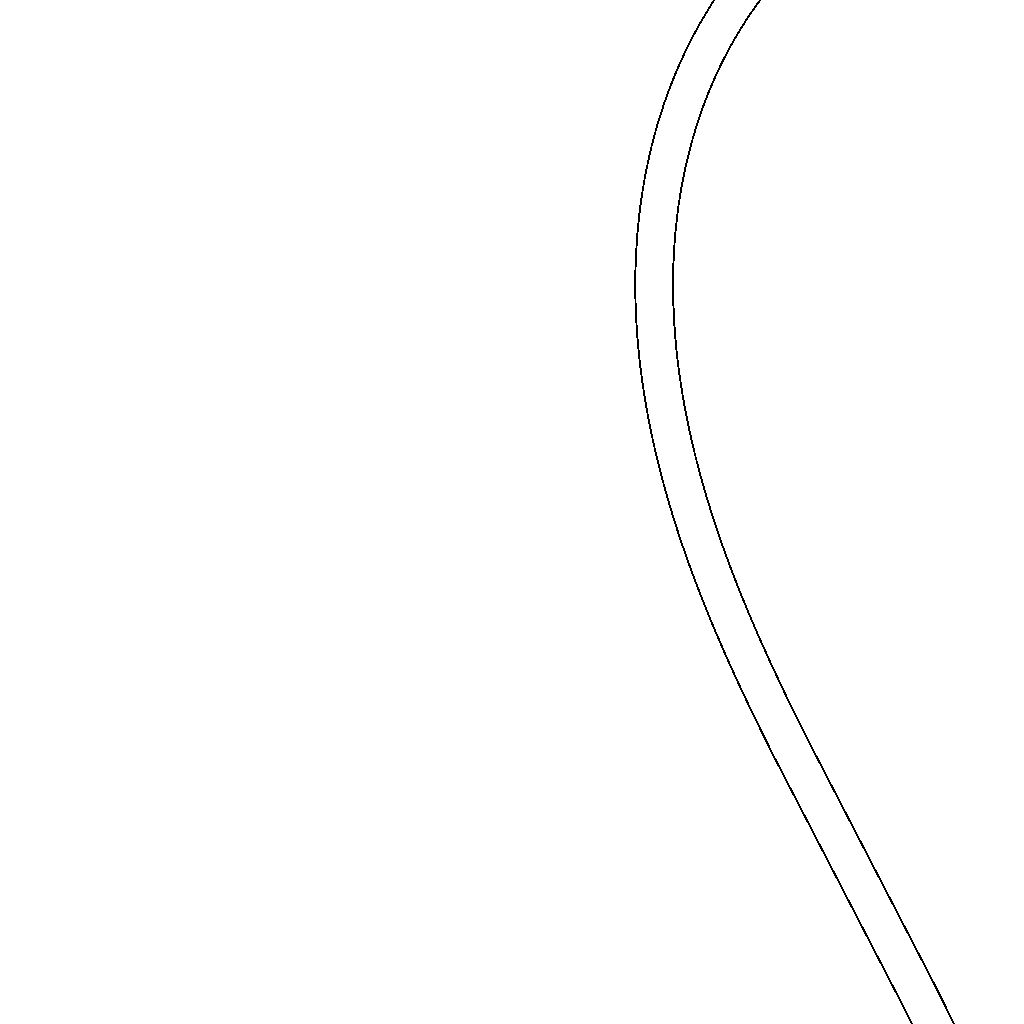}} &
		\fbox{\includegraphics[width=0.18\linewidth]{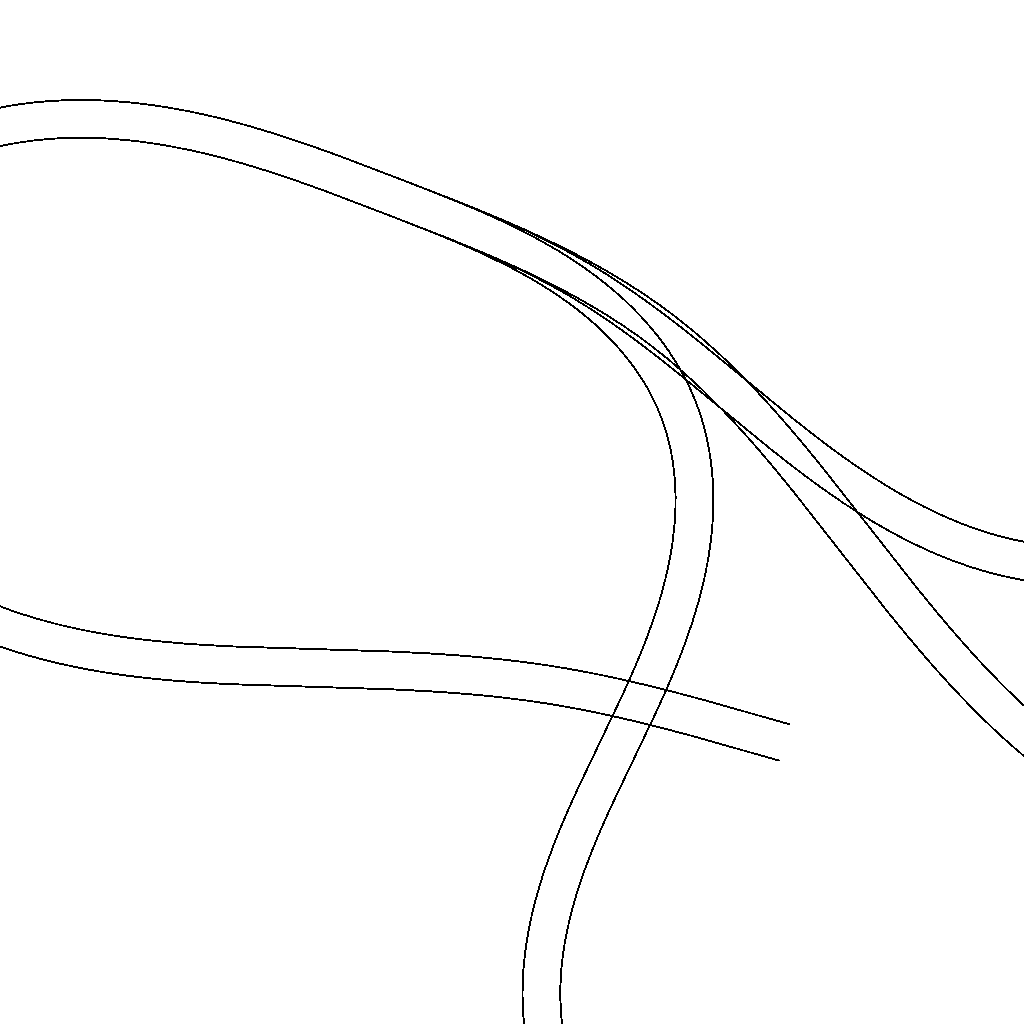}} &
		\fbox{\includegraphics[width=0.18\linewidth]{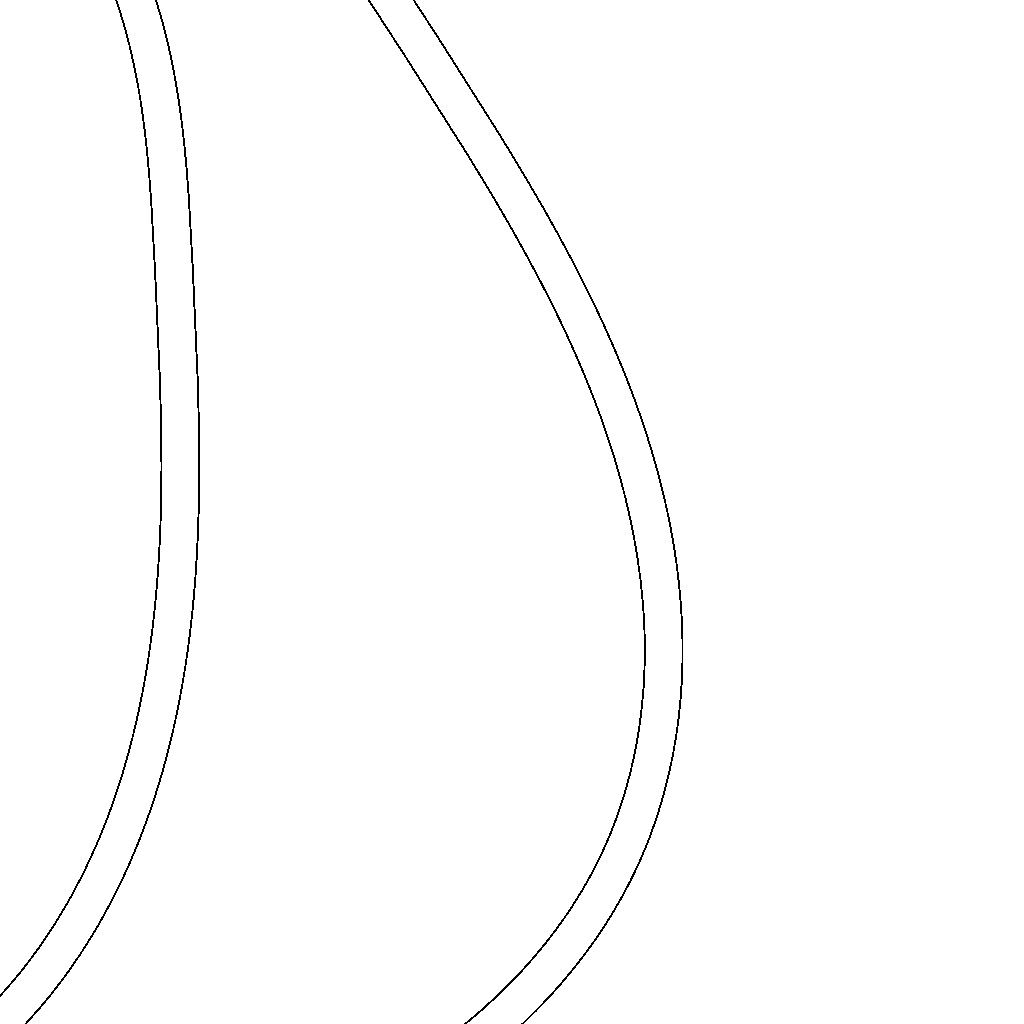}} \\[\smallskipamount]
		\fbox{\includegraphics[width=0.18\linewidth]{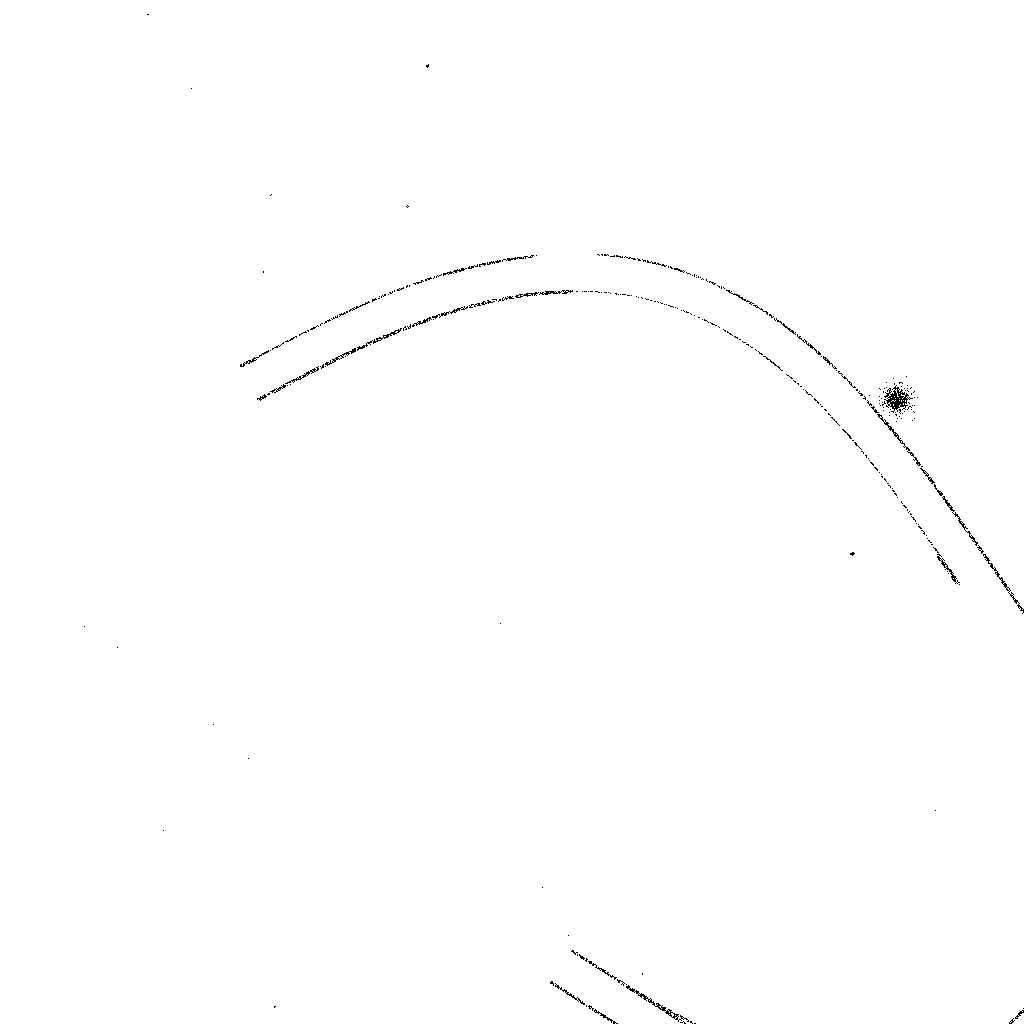}} &
		\fbox{\includegraphics[width=0.18\linewidth]{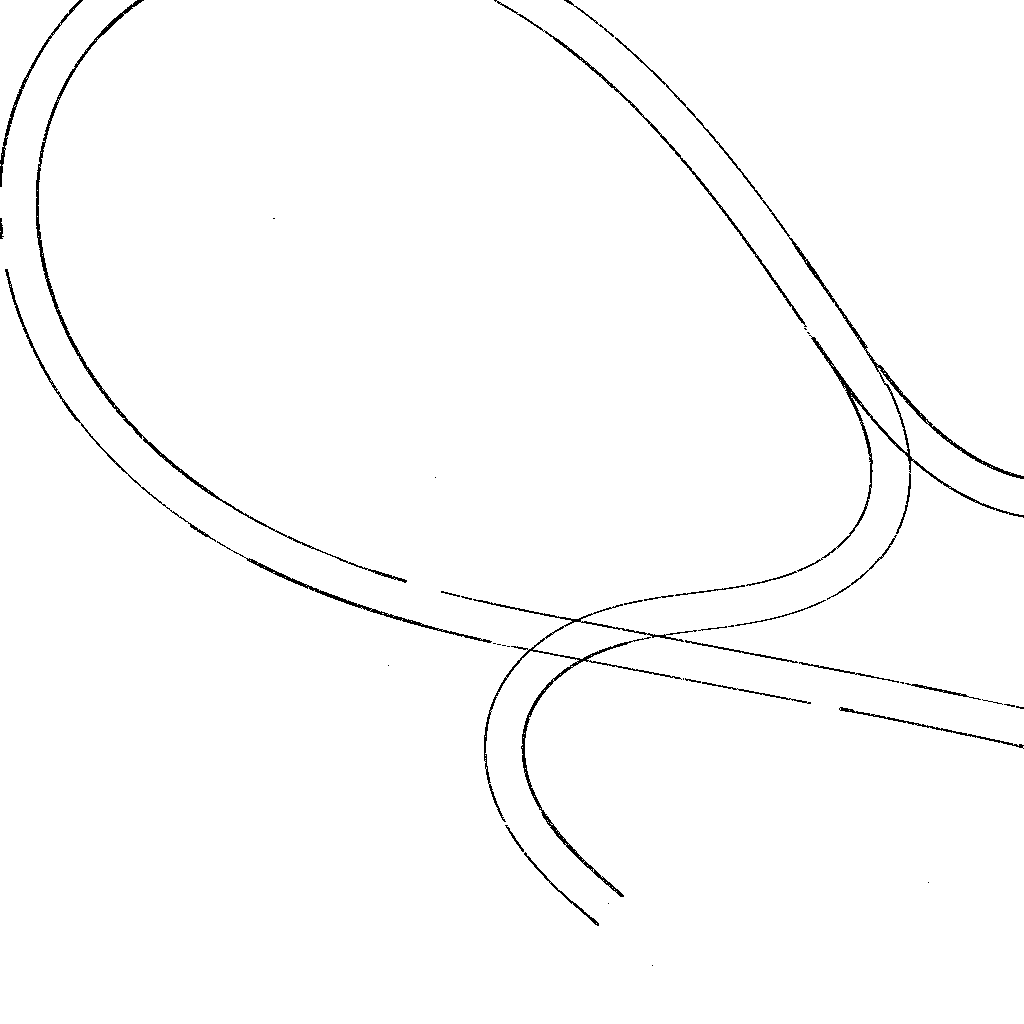}} &
		\fbox{\includegraphics[width=0.18\linewidth]{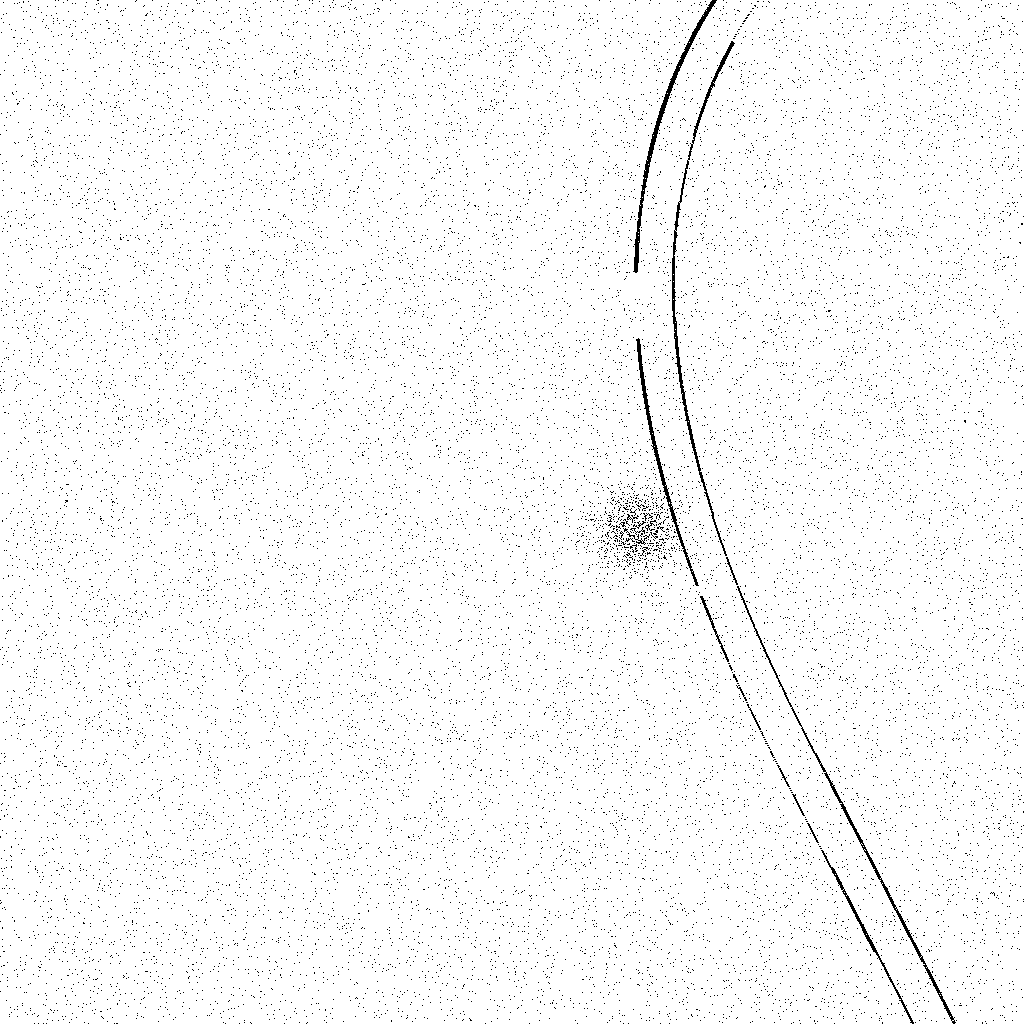}} &
		\fbox{\includegraphics[width=0.18\linewidth]{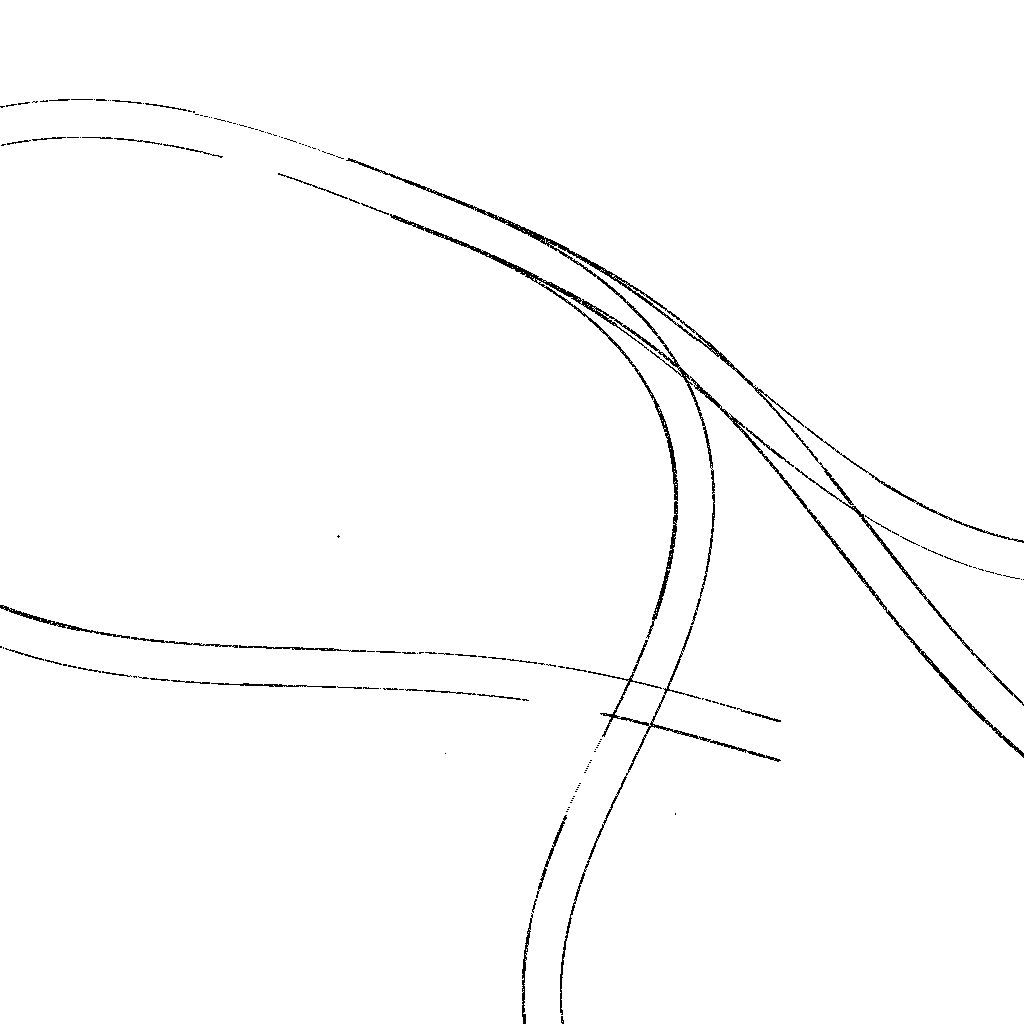}} &
		\fbox{\includegraphics[width=0.18\linewidth]{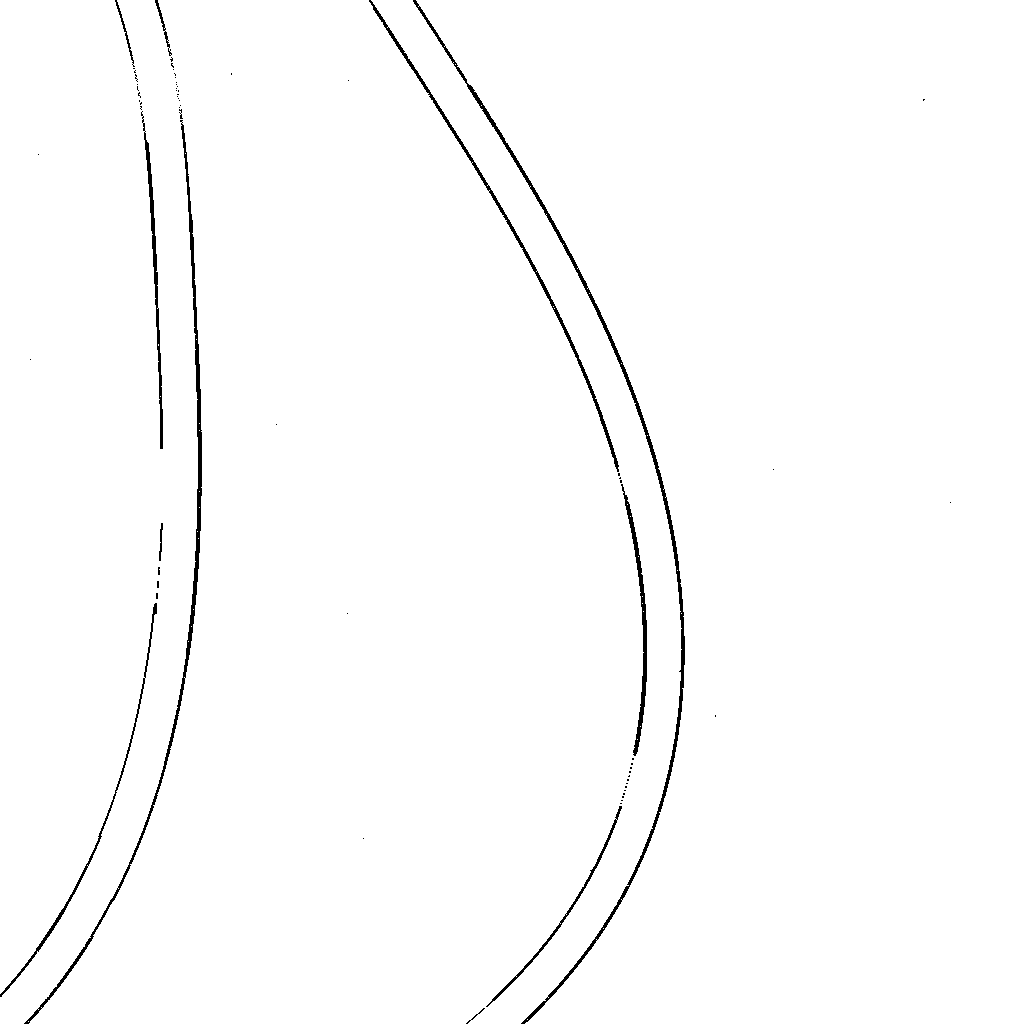}}
	\end{tabular}
	\caption{Synthetic training examples. The top row shows the original noiseless track rasters; the bottom row shows the corresponding distorted rasters with added noise.  Each column pairs the same layout before and after corruption.}
	\label{fig:synthetic_data_examples}
\end{figure}

Figure~\ref{fig:synthetic_data_examples} shows five representative synthetic track layouts rendered with the following U.S.~132~RE parameters: gauge~\SI{1.435}{\metre}, railhead thickness~\SI{76.2}{\milli\metre}, and raster cell size~\SI{4}{\centi\metre}.  At this resolution each rail spans roughly two pixels, giving the network just enough detail to capture sub-pixel offsets while keeping the rasters compact.  The generator produces a broad range of geometries\textemdash broad curves, tight bends, transitions, and split junctions\textemdash which are visible in the top row. Each clean raster is paired with a corrupted counterpart created by combining rail shifts, missing segments, local thickness variations, pixel flips, and random blob artifacts.  Training on this aggressively distorted set encourages the model to rely on track-level geometry\textemdash two continuous, parallel rails at constant gauge\textemdash rather than superficial pixel patterns.

\setlength{\fboxsep}{0pt}
\setlength{\fboxrule}{0.5pt}

Validation of neural network performance was conducted primarily using synthetic datasets, as substantial volumes of synthetic data allowed comprehensive training and comparative analyses of different neural network architectures.

\subsection{Simulation of Railroad Network and Rail Track Centerlines}

Two strategies were implemented to generate the railroad network. The first strategy randomly selects parameters for railroad splits and determines whether segments will be straight or curved. The second strategy involves generating random points with randomly assigned directions within a defined rectangular area, simulating varied railroad layouts. These points are subsequently randomly connected using clothoid curves, ensuring continuous and smooth transitions between rail track segments. This process utilizes a precomputed table containing combinations of clothoid curves for optimal point-to-point connectivity.

This approach generates both realistic and unrealistic railroad designs. The primary objective is to achieve sufficient variability to encompass all realistic layouts, accepting the inclusion of numerous unrealistic configurations as a necessary trade-off.

To derive the individual rail track centerlines, the computed railroad centerline must be shifted by the half-gauge distance plus the half-rail thickness in both directions.

\subsection{Track Distortion and Noise Addition}
Subsequent to generating the idealized centerlines, the data underwent several distortion processes to mimic realistic track conditions. Initial distortion was applied by perturbing rail geometries and occasionally omitting segments, simulating potential data-capture errors.

The distorted rail lines were rasterized into grid-based representations, facilitating the introduction of pixel-level noise. This additional noise included random pixel flips based on predefined probabilities. Additional randomized blob-like artifacts, varying in size and number, were introduced to further represent diverse and realistic noise scenarios. Noise probabilities and blob characteristics were randomly selected to ensure broad coverage of potential noise scenarios, ranging from large-scale disturbances to minor, localized irregularities.

\section{Neural Network Architecture\label{sec:NeuralNetworkArchitecture}}

The architecture is built upon a full-resolution, fully convolutional recurrent neural network (RNN) design. Each recurrent block within this structure consists of a fully convolutional neural network employing dilated convolutions to effectively capture spatial details and multiscale contextual information.

While conceptually similar to the U-Net architecture \cite{UNet, UNetGithub} initially introduced in \cite{UNet2015}, our method differs fundamentally by replacing max-pooling layers with dilation layers. This critical modification ensures that full spatial resolution is maintained throughout the network, preventing the loss of detail commonly encountered with downsampling methods. Our empirical observations indicate that conventional downsampling architectures are insufficient for reconstructing railroad tracks due to uneven pixel processing, which introduces artifacts problematic for vectorization tasks. Addressing this challenge, our approach integrates dilated convolutional layers (also referred to as \textit{atrous convolutions}) \cite{DilationLayer}, which produce high-quality, uniformly processed pixel representations essential for accurate rail track centerline reconstruction.

Similar to the ACNN~\cite{ACNN, ACNNGithub}, our architecture operates at full resolution to prevent these common artifacts. Although maintaining full resolution does increase computational demands due to the absence of downsampling, it does not inherently require more parameters. We mitigate the additional computational burden by decreasing the number of channels, thus removing the necessity to aggregate information from lower-resolution feature maps.

Through empirical experimentation, we found that skip connections\textemdash widely used in U-Net for transferring detailed layer information\textemdash were not beneficial in our context. Since our neural network maintains full resolution, these skip connections\textemdash by continuously reintroducing raw, unprocessed features\textemdash unintentionally restricted the network's capacity to derive more complex and abstract features.

Our findings align with those presented in the ACNN architecture, which similarly identified the absence of skip connections as being advantageous for detailed pixel-wise prediction tasks. Furthermore, dilated convolutions allow our network to substantially increase its receptive field without a significant increase in parameters, effectively capturing broader spatial contexts while preserving computational efficiency.

The proposed architecture, FRPDF, integrates progressive fusion via recurrent dilated convolutions. This design effectively merges multiscale information without reducing input resolution, making it particularly well-suited for precise pixel-level tasks such as rail track extraction.

\begin{figure}[!htbp]
	\centering
	\begin{tikzpicture}[
		node distance=0.7cm and 0.7cm,
		every node/.style={font=\small},
		layer/.style={rectangle, draw, rounded corners, minimum width=2.8cm, minimum height=1.0cm, align=center, fill=blue!10},
		combine/.style={rectangle, draw, dashed, minimum width=2.8cm, minimum height=1.0cm, align=center, fill=orange!10},
		input/.style={rectangle, draw, minimum width=2.8cm, minimum height=1.0cm, align=center, fill=green!10},
		output/.style={rectangle, draw, minimum width=2.8cm, minimum height=1.0cm, align=center, fill=red!10},
		pdf/.style={rectangle, draw, minimum width=2.8cm, minimum height=1.0cm, align=center, fill=yellow!10},
		arrow/.style={-Latex, thick},
		skip/.style={-Latex, thick, dashed, red}
		]
		
		\node (x_in) {in};
		
		\node[layer, right=0.5cm of x_in] (conv) {
			Conv2d\\
			in\_channels: $C_\mathrm{in}$\\
			out\_channels: $C_\mathrm{out}$\\
			kernel\_radius: $r$\\
			padding: $r \cdot d$\\
			dilation: $d$\\
			no bias
		};
		\node[layer, right=0.5cm of conv] (bn) {
			BatchNorm2d\\
			channels: $C_\mathrm{out}$
		};
		\node[layer, right=0.5cm of bn] (relu) {
			ReLU\\
			channels: $C_\mathrm{out}$
		};
		
		\node[right=0.5cm of relu] (x_out) {out};
		
		\draw[arrow] (x_in) -- (conv);
		\draw[arrow] (conv) -- (bn);
		\draw[arrow] (bn) -- (relu);
		\draw[arrow] (relu) -- (x_out);
		
		\node[draw, dashed, fit=(conv)(bn)(relu), label=above:{CustomConv2d Block}] (block) {};
	\end{tikzpicture}
	\caption{A custom 2D convolution block. The Conv2d node has $C_{\mathrm{in}}$ input channels and $C_{\mathrm{out}}$ output channels, kernel radius $r$, padding $r \cdot d$, dilation $d$, and no bias. The subsequent BatchNorm2d and ReLU use $C_{\mathrm{out}}$ channels.}
	\label{fig:customconv_block_horizontal}
\end{figure}
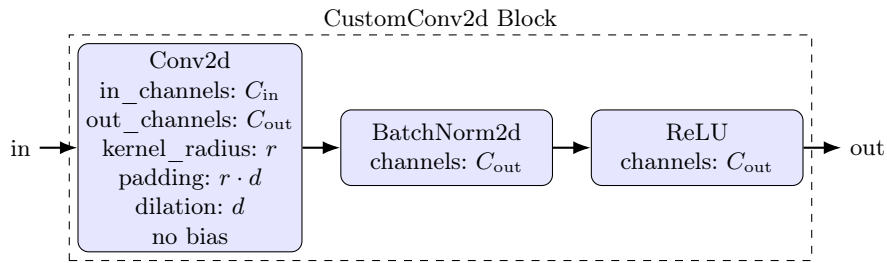

Throughout this section, we use \textit{kernel radius} $r$ to denote the integer half-width of a square convolution kernel; thus, the corresponding kernel size is $(2r + 1) \times (2r + 1)$. Figure~\ref{fig:customconv_block_horizontal} shows the custom convolution block, \textit{CustomConv2d}, from \cite{UNetGithub}, which comprises a \textit{Conv2d} layer configured with $C_{\mathrm{in}}$ input channels and $C_{\mathrm{out}}$ output channels, a kernel radius $r$, padding $r \cdot d$, dilation factor $d$, and no bias. The output is subsequently processed by a \textit{BatchNorm2d} layer \cite{BatchNormalization} and activated with \textit{ReLU}, both operating on $C_{\mathrm{out}}$ channels.

\begin{figure}[!htbp]
	\centering
	\begin{tikzpicture}[
		node distance=0.7cm and 0.7cm,
		every node/.style={font=\small},
		layer/.style={rectangle, draw, rounded corners, minimum width=2.8cm, minimum height=1.0cm, align=center, fill=blue!10},
		combine/.style={rectangle, draw, dashed, minimum width=2.8cm, minimum height=1.0cm, align=center, fill=orange!10},
		input/.style={rectangle, draw, minimum width=2.8cm, minimum height=1.0cm, align=center, fill=green!10},
		output/.style={rectangle, draw, minimum width=2.8cm, minimum height=1.0cm, align=center, fill=red!10},
		pdf/.style={rectangle, draw, minimum width=2.8cm, minimum height=1.0cm, align=center, fill=yellow!10},
		arrow/.style={-Latex, thick},
		skip/.style={-Latex, thick, dashed, red}
		]

		\node (x_in) {in};

		\node[layer, right=0.5cm of x_in] (blockA) {
			CustomConv2d\\
			in\_channels: $C_\mathrm{in}$\\
			out\_channels: $C_\mathrm{out}$\\
			kernel\_radius: $r$\\
			dilation: $d$
		};
		\node[layer, right=0.5cm of blockA] (blockB) {
			CustomConv2d\\
			in\_channels: $C_\mathrm{out}$\\
			out\_channels: $C_\mathrm{out}$\\
			kernel\_radius: $r$\\
			dilation: $d$
		};
		
		\node[right=0.5cm of blockB] (x_out) {out};
		
		\node[draw, dashed, fit=(blockA)(blockB), label=above:{StackedConv2d}] (stacked) {};
		
		\draw[arrow] (x_in) -- (blockA);
		\draw[arrow] (blockA) -- (blockB);
		\draw[arrow] (blockB) -- (x_out);
	\end{tikzpicture}
	\caption{A stacked arrangement of two consecutive \textit{CustomConv2d} blocks. Here, 
		$C_\mathrm{in}$ and $C_\mathrm{out}$ denote the number of input and output 
		feature channels, respectively; $r$ is the kernel radius; and $d$ is 
		the dilation factor. The first block uses $C_\mathrm{in}$ as input channels 
		and $C_\mathrm{out}$ as output channels, while the second block reuses 
		$C_\mathrm{out}$ for both its input and output.}
	\label{fig:StackedConv2d}
\end{figure}
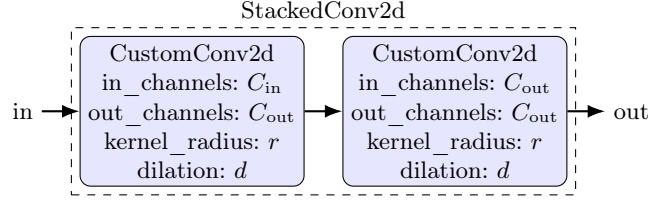

Figure~\ref{fig:StackedConv2d} shows the stacked convolution module, \textit{StackedConv2d}, from \cite{UNetGithub}, which consists of two sequential custom convolution blocks. The first block processes the input with $C_{\mathrm{in}}$ channels and produces $C_{\mathrm{out}}$ channels, while the second block uses $C_{\mathrm{out}}$ channels for both its input and output. Both convolution blocks use kernel radius $r$ and dilation factor $d$.

\begin{figure}[!htbp]
	\centering
	\begin{tikzpicture}[
		node distance=1.0cm and 1.0cm,
		every node/.style={font=\small},
		layer/.style={rectangle, draw, rounded corners, minimum width=3.0cm, minimum height=1.2cm, align=center, fill=blue!10},
		arrow/.style={-Latex, thick},
		skip/.style={-Latex, thick, dashed, red}
		]
		
		\node (x_in) {in};
		
		\node[layer, right=0.5cm of x_in] (layerA) {
			CustomConv2d\\
			in\_channels: $c$\\
			out\_channels: $c$\\
			kernel\_radius: $r$\\
			dilation: $d$
		};
		
		\node[draw, circle, right=1cm of layerA] (concat) {Concat};
		
		\node[layer, right=of concat] (layerB) {
			CustomConv2d\\
			in\_channels: $2 c$\\
			out\_channels: $c$\\
			kernel\_radius: 0\\
			dilation: 1
		};
		
		\node[right=0.5cm of layerB] (x_out) {out};
		
		\draw[arrow] (x_in) -- (layerA);
		\draw[arrow] (layerA) -- (concat);
		\draw[skip]  (x_in) |- ++(0,-1.2) -| (concat);
		\draw[arrow] (concat) -- (layerB);
		\draw[arrow] (layerB) -- (x_out);
		
		\node[draw, dashed, fit=(layerA)(concat)(layerB),
		label=above:{ProgressiveCombine2d}] (block) {};
		
	\end{tikzpicture}
	\caption{A \textit{ProgressiveCombine2d} block. The first \textit{CustomConv2d} has \(c\) 
		input/output channels. Its output is concatenated with the original input (via a dashed skip) 
		to form \(2 c\) channels, which the second \textit{CustomConv2d} then reduces back to \(c\) channels.}
	\label{fig:progressive_combine_block}
\end{figure}
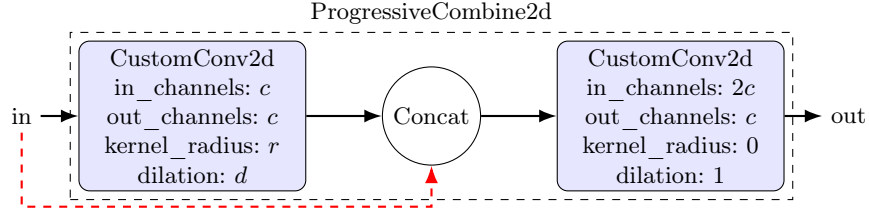

Figure~\ref{fig:progressive_combine_block} shows a progressively combined block, \textit{ProgressiveCombine2d}. In this block, an initial \textit{CustomConv2d} layer with \(c\) channels is applied to the input. Its output is concatenated with the original input via a skip connection, forming 
\(2 c\) channels that are then reduced back to \(c\) channels by a subsequent \textit{CustomConv2d} layer.

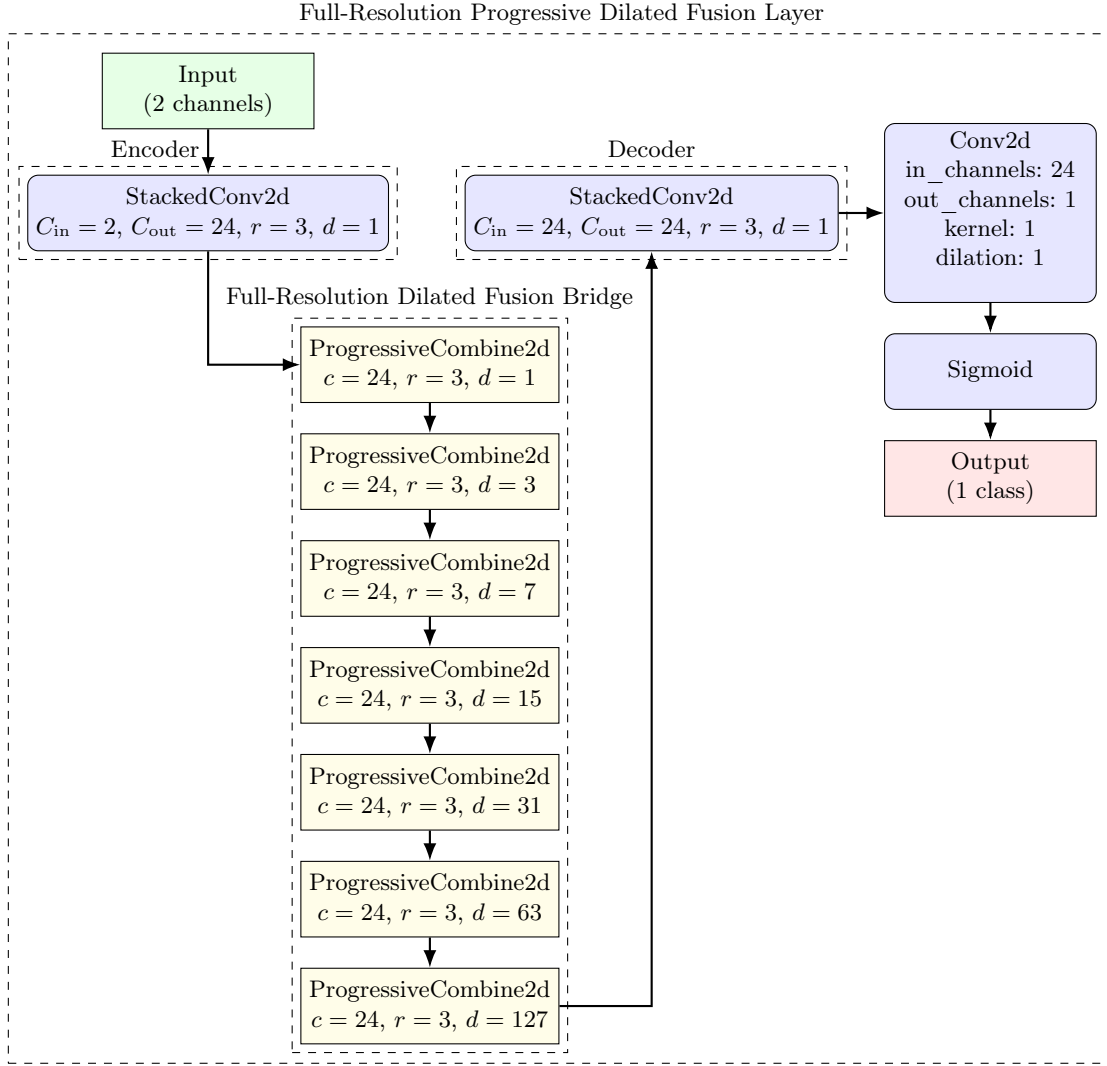
\begin{figure}[!htbp]
	\centering
	\begin{tikzpicture}[
		node distance=0.7cm and 0.7cm,
		every node/.style={font=\small},
		layer/.style={rectangle, draw, rounded corners, minimum width=2.8cm, minimum height=1.0cm, align=center, fill=blue!10},
		combine/.style={rectangle, draw, dashed, minimum width=2.8cm, minimum height=1.0cm, align=center, fill=orange!10},
		input/.style={rectangle, draw, minimum width=2.8cm, minimum height=1.0cm, align=center, fill=green!10},
		output/.style={rectangle, draw, minimum width=2.8cm, minimum height=1.0cm, align=center, fill=red!10},
		pdf/.style={rectangle, draw, minimum width=2.8cm, minimum height=1.0cm, align=center, fill=yellow!10},
		arrow/.style={-Latex, thick},
		skip/.style={-Latex, thick, dashed, red}
		]
		
		\node[input] (input) {Input\\(2 channels)};
		
		\node[layer, below=0.6cm of input] (enc1) {StackedConv2d\\$C_\mathrm{in}=2$, $C_\mathrm{out}=24$, $r=3$, $d=1$};
		
		\node[draw, dashed, fit=(enc1)(enc1), label={[above, anchor=south east]Encoder}] {};		
		
		\node[layer, right=1.0cm of enc1] (dec1) {StackedConv2d\\$C_\mathrm{in}=24$, $C_\mathrm{out}=24$, $r=3$, $d=1$};
		
		\node[draw, dashed, fit=(dec1)(dec1), label={[above, anchor=south]Decoder}] {};
		
		\draw[arrow] (input) -- (enc1);
		
		\coordinate (pdfStart) at ($ (enc1.south)!0.5!(dec1.south) + (0,-1.5cm) $);
		\node[pdf] (pdf1) at (pdfStart) {ProgressiveCombine2d\\$c=24$, $r=3$, $d=1$};
		\node[pdf, below=0.4cm of pdf1] (pdf2) {ProgressiveCombine2d\\$c=24$, $r=3$, $d=3$};
		\node[pdf, below=0.4cm of pdf2] (pdf3) {ProgressiveCombine2d\\$c=24$, $r=3$, $d=7$};
		\node[pdf, below=0.4cm of pdf3] (pdf4) {ProgressiveCombine2d\\$c=24$, $r=3$, $d=15$};
		\node[pdf, below=0.4cm of pdf4] (pdf5) {ProgressiveCombine2d\\$c=24$, $r=3$, $d=31$};
		\node[pdf, below=0.4cm of pdf5] (pdf6) {ProgressiveCombine2d\\$c=24$, $r=3$, $d=63$};
		\node[pdf, below=0.4cm of pdf6] (pdf7) {ProgressiveCombine2d\\$c=24$, $r=3$, $d=127$};
		
		\node[draw, dashed, fit=(pdf1)(pdf7), label=above:{Full-Resolution Dilated Fusion Bridge}] {};
		
		\draw[arrow] (enc1.south) -- ++(0,0) |- (pdf1.west);
		
		\draw[arrow] (pdf1)  -- (pdf2);
		\draw[arrow] (pdf2)  -- (pdf3);
		\draw[arrow] (pdf3)  -- (pdf4);
		\draw[arrow] (pdf4)  -- (pdf5);
		\draw[arrow] (pdf5)  -- (pdf6);
		\draw[arrow] (pdf6)  -- (pdf7);
		
		

		\node[layer, right=0.6cm of dec1] (finalConv) {			
			Conv2d\\
			in\_channels: $24$\\
			out\_channels: $1$\\
			kernel: $1$\\
			dilation: $1$\\
		};
		\node[layer, below=0.4cm of finalConv] (sigmoid) {Sigmoid};
		
		\node[output, below=0.4cm of sigmoid] (output) {Output\\(1 class)};
		
		\draw[arrow] (dec1) -- (finalConv);
		\draw[arrow] (finalConv) -- (sigmoid);
		\draw[arrow] (sigmoid) -- (output);

		\draw[arrow] (pdf7.east) -- ++(0,0) -| (dec1.south);
		
		
		\node[draw, dashed, fit=(input)(enc1)(enc1)(dec1)(dec1)(pdf7)(output), inner sep = 0.25cm, label=above:{Full-Resolution Progressive Dilated Fusion Layer}] {};
		
	\end{tikzpicture}
	\caption
	{
		The network begins with a channel-expansion front end\textemdash informally called an \QuotationMarks{encoder}\textemdash that projects the two-channel input into $24$ feature maps while preserving full spatial resolution. These $24$-channel maps feed into the Full-Resolution Dilated Fusion bridge, which comprises seven dilated-convolution blocks with dilation rates $ 2^n - 1, n = \overline{1\dots7}$ (i.e. $\left\{1,3,7,15,31,63,127\right\}$), to sequentially aggregate multiscale context without any spatial down- or up-sampling. A symmetric $24$-channel back end\textemdash serving as a \QuotationMarks{decoder} only in the sense of mirroring the front-end convolutions\textemdash further processes these features but retains the same channel count. Finally, a $1\times1$ convolution collapses the $24$ channels to a single feature map, and a sigmoid activation produces the per-pixel single-class probability output.
	}
	\label{fig:nn_recurrent_block}
\end{figure}

FRPDF architecture is shown in Figure~\ref{fig:nn_recurrent_block}. The architecture leverages dilated convolutions at varying dilation rates within its central fusion bridge to efficiently integrate multiscale contextual information without compromising spatial resolution. These dilated convolutions effectively expand the receptive field, allowing the network to consider broader spatial contexts and maintain high-resolution feature maps simultaneously.

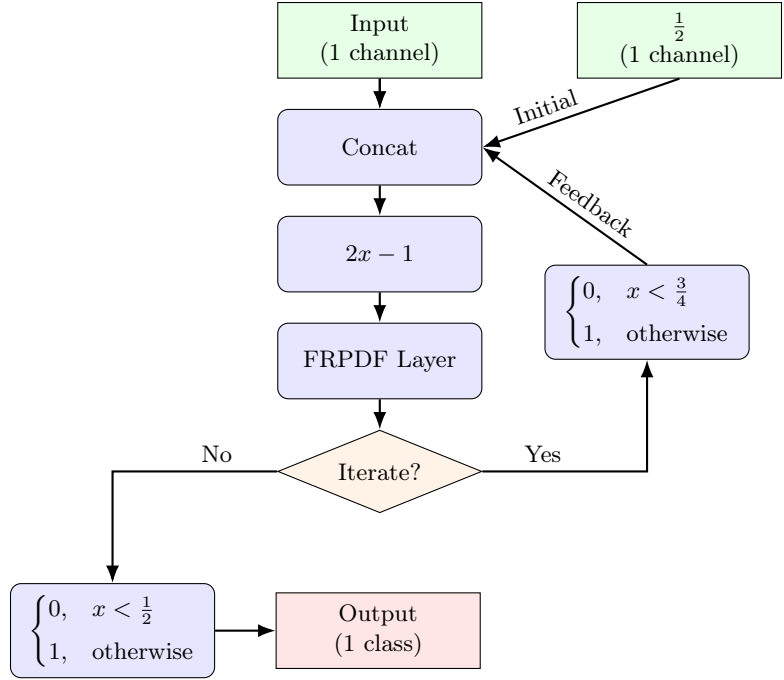
\begin{figure}[!htbp]
	\centering
	\begin{tikzpicture}[
		node distance=1.6cm and 2.5cm,
		every node/.style={font=\small},
		layer/.style={rectangle, draw, rounded corners, minimum width=2.7cm, minimum height=1cm, align=center, fill=blue!10},
		input/.style={rectangle, draw, minimum width=2.7cm, minimum height=1cm, align=center, fill=green!10},
		output/.style={rectangle, draw, minimum width=2.7cm, minimum height=1cm, align=center, fill=red!10},
		decision/.style={diamond, draw, aspect=2, minimum width=2.7cm, minimum height=1cm, align=center, fill=orange!10},
		arrow/.style={-Latex, thick}
		]
		
		\node[input] (xIn) {Input\\(1 channel)};
		
		\node[layer, below=0.4cm of xIn] (concat) {Concat};
		
		\node[input, right=1.25cm of xIn] (initVal) {\(\tfrac12\)\\(1 channel)};
		
		\node[layer, below=0.4cm of concat] (arith) {\(2x - 1\)};
		\node[layer, below=0.4cm of arith] (hdf) {FRPDF Layer};
		
		\node[decision, below=0.4cm of hdf] (decision) {Iterate?};
		
		\node[layer, below left=1.2cm and 1.5cm of decision] (thresh12) {
			\(\displaystyle
			\begin{cases}
				0, & x < \tfrac12\\[1mm]
				1, & \text{otherwise}
			\end{cases}
			\)};
		\node[layer, above right=1.2cm and 1.5cm of decision] (thresh34) {
			\(\displaystyle
			\begin{cases}
				0, & x < \tfrac34\\[1mm]
				1, & \text{otherwise}
			\end{cases}
			\)};
		
		\node[output, right=0.8cm of thresh12] (xOut) {Output\\(1 class)};
		
		\draw[arrow] (initVal.south) -- node[pos=0.65, sloped, above]{Initial} (concat.east);
		\draw[arrow] (xIn.south) -- (concat.north);
		\draw[arrow] (xIn) -- (concat);
		
		\draw[arrow] (concat) -- (arith);
		\draw[arrow] (arith) -- (hdf);
		\draw[arrow] (hdf) -- (decision);
		
		\draw[arrow] (decision.west) -- ++(-0.8,0)node[above]{No} -| (thresh12.north);
		\draw[arrow] (decision.east) -- ++(0.8,0)node[above]{Yes} -| (thresh34.south);
		
		\draw[arrow] (thresh12) -- (xOut);
		
		\draw[arrow] (thresh34.north) -- node[pos=0.4, sloped, above]{Feedback} (concat.east);
		
	\end{tikzpicture}
	\caption
	{
		A recurrent architecture with a decision diamond after the FRPDF layer. From the diamond block, the \textbf{No} arrow (exiting left and down) proceeds to the \(\tfrac12\) threshold and then to the final output, and the \textbf{Yes} arrow (exiting right and up) loops recurrently via the \(\tfrac34\) threshold, which feeds back to the Concat block.
	}
	\label{fig:nn_architecture_simplified}
\end{figure}

The FRPDF layer is used as a recurrent layer, as shown in Figure~\ref{fig:nn_architecture_simplified}. In our experiments, we found that performing two iterations of the FRPDF processing significantly improves the quality of the results, especially when handling datasets containing larger gaps or noisy regions. Additional iterations beyond two further enhance results, but the improvement is less dramatic compared to the benefit gained from the initial two iterations. Although this paper uses a fixed number of iterations, the criteria for deciding the optimal number of iterations can be designed dynamically to better adapt to varying input data conditions.

\section{Quality Evaluation\label{sec:QualityEvaluation}}

Our evaluation focuses on rail track extraction performance using confusion-matrix metrics\textemdash false negatives (FN), false positives (FP), and true positives (TP)\textemdash but not true negatives (TN) since they predominantly cover nonrail areas and add little insight. We have opted \emph{not} to include a direct ACNN baseline here\textemdash we found that a truly fair parameter- and capacity-matched comparison was infeasible without extensive reengineering beyond the scope of this work\textemdash so we instead report FRPDF's stand-alone performance in Table~\ref{tab:comparison}.

\begin{table}[htb]
	\centering
	\caption{FRPDF performance was evaluated on $1024\times1024$ test images (batch size of 2) using a Dell workstation with an Intel Xeon W-2265 (\SI{3.5}{\giga\hertz}, 12-core, Hyper-Threading), \SI{64}{\giga\byte} RAM, and an NVIDIA RTX A6000 GPU. Metrics: precision, recall, and F1 score. Resource measures: parameter count, peak GPU memory, and per-image inference time.}
	\label{tab:comparison}
	\begin{tabular}{|l|r|r|r|c|c|c|r|c|c|}
		\hline
		Model & FN & FP & TP & Precision & Recall & F1 & Parameters & GPU Memory & Time \\
		\hline
		FRPDF & 214{,}014 & 875{,}941 & 3{,}716{,}533 & 0.809 & 0.946 & 0.872 & 293{,}545 & \SI[parse-numbers=false]{1{,}554}{\mega\byte} & \SI{98}{\milli\second} \\
		\hline
	\end{tabular}
\end{table}

\section{Applying the Neural Network to Real Data\label{sec:ApplyingTheNeuralNetworkToRealData}}
The FRPDF networks were trained entirely on synthetic data covering common European and U.S. track standards, along with a sparse set of additional pairs of track gauge and railhead thickness to improve generalization.

We first select the model whose training parameters have the ratio closest to the target track gauge and railhead thickness. We scale the input lidar points accordingly, rasterize them, and then pass them through the network. To remove edge effects, an overlapping tiling scheme is used. This procedure allows several pretrained models to cover a wide range of real-world rail configurations without retraining.

\section{Vectorization\label{sec:Vectorization}}
After inference, the FRPDF network yields a cleaned binary raster in which most noise is suppressed, small gaps are bridged, and the rail profile is enforced. Minor artifacts nevertheless remain, so direct vectorization would still introduce spurious branches. We therefore apply a morphological \emph{closing} to fill small holes and bridge short gaps. The resultant raster is vectorized using veinerization algorithm after applying \emph{parity-fix} transformation, described in~\cite{VectorizationAndParityErrors}. A comparable workflow is provided by the ArcScan~\cite{ArcScanManual, ArcScanDesktopHelp} extension for ArcGIS.  

\subsection{Postprocessing of Vectorized Geometry}
The polylines obtained from vectorization are smoothed using the algorithm described in~\cite{Smoothing, SmoothingNetwork}. Each lidar point classified as \textit{rail} is orthogonally projected onto the nearest smoothed polyline while preserving its original $z$-coordinate, and the projected points are ordered along the rail. The extremal-boundary algorithm described in \cite{EfficientComputationOfTheDirectionalExtremalBoundaryOfAUnionOfEqualRadiusCircles} is applied and only those points lying on the railhead top surface are retained. Finally, the $z$-profile is smoothed with the same one-dimensional smoothing algorithm~\cite{Smoothing, SmoothingNetwork}, yielding geometrically smooth 3D polylines representing the centerlines of the top of the rails.

\section{Railroad Track Centerline Reconstruction\label{sec:RailroadTrackCenterlineReconstruction}}

\subsection{Matching Pairs of Polylines}

The matching of polyline pairs is based on DTW~\cite{DynamicTimeWarpingAlgorithmReview}. Given two polylines, \(P_I\) and \(P_J\), the algorithm identifies common subpolylines within the specified distance tolerance \(T\).

\subsubsection*{Element Pairing}
To reduce complexity, the algorithm focuses on point-to-segment (or segment-to-point) matches. A pair is deemed valid if the minimum distance between the point and the segment is less than \(T\). In a point-to-segment match, two point-to-point relationships are implicitly formed: one between the point and the segment's start vertex, and another between the point and the segment's end vertex. Because polylines can overlap in multiple areas, numerous candidate pairs may be generated.

\subsubsection*{Sorting Candidates}
All valid pairs are sorted by their positions along both polylines, ensuring that each unique pair is considered only once. The sorting enforces the rule that no pair \((p_1, q_1)\) follows another pair \((p_2, q_2)\) if \(p_2 < p_1\) but \(q_2 \leq q_1\), or vice versa. This consistent ordering is crucial for subsequent steps. Sorting by the indices of one polyline and then by the other allows for efficient binary searches that verify pair existence and update the best match. Each pair stores its longest match length, the best (lowest) penalty found so far, and a reference to its predecessor in the match sequence.

\subsubsection*{Continuation Check and Cluster Formation}
For each pair, the algorithm checks whether it can be extended by advancing one step along \(P_I\) or \(P_J\). If a valid continuation is found, the clusters containing these pairs are merged. The algorithm tracks the total length of the matched subpolylines; if extending the match results in a length greater than the current maximum\textemdash or matches the current maximum while having a lower penalty\textemdash the match is updated. The penalty is computed as the integral of the squared distance between a point on one polyline and its corresponding segment on the other, as described in \secrefplain{sec:IntegralSegmentToPoint}. By maintaining a reference to the previous pair, the algorithm can reconstruct the entire matched subpolylines.

\subsubsection*{Selection of Final Matches}
After all candidate pairs have been processed, the algorithm selects matches in descending order of their subpolyline length. It skips any pairs belonging to clusters already in use, to avoid overlap. Finally, the matching subpolylines for each selected match are reconstructed using references to previous pairs.

\section{Example\label{sec:Example}}

The proposed algorithm has been implemented in the Extract Rails From Point Cloud tool in ArcGIS~Pro~3.4 \cite{ReferenceExtractRailsFromPointCloud}. In this section, we use the tool to extract rail geometries from the rail points classified in \secref{sec:DataAndRailClassification}.

The LAS file consists of $75,499$ classified rail points (class $10$). The classification of rail points took \SI{42}{\second}. The dataset spans three railway tracks, each about \SI{160}{\metre} in length and exhibiting varying point densities, discontinuities, and noise, with a small number of misclassified rail points located along or between the rails.

Parameters are configured to match the physical characteristics of the rail tracks. A track gauge of \SI{1}{\metre} and a rail thickness of \SI{61.91}{\milli\metre} are specified to reflect the actual rail geometry. Detected rail points are rasterized and processed by the FRPDF network to produce a binary raster representation. This raster is subsequently vectorized to produce preliminary rail line features. The classified rail points are used to transfer $z$-values. The extracted rail lines are then refined using the smoothing and simplification algorithms. Finally, rail centerlines are derived from the pairs of matched rail polylines.

\begin{figure}[!htbp]
	\centering
	\includegraphics[width=\linewidth]{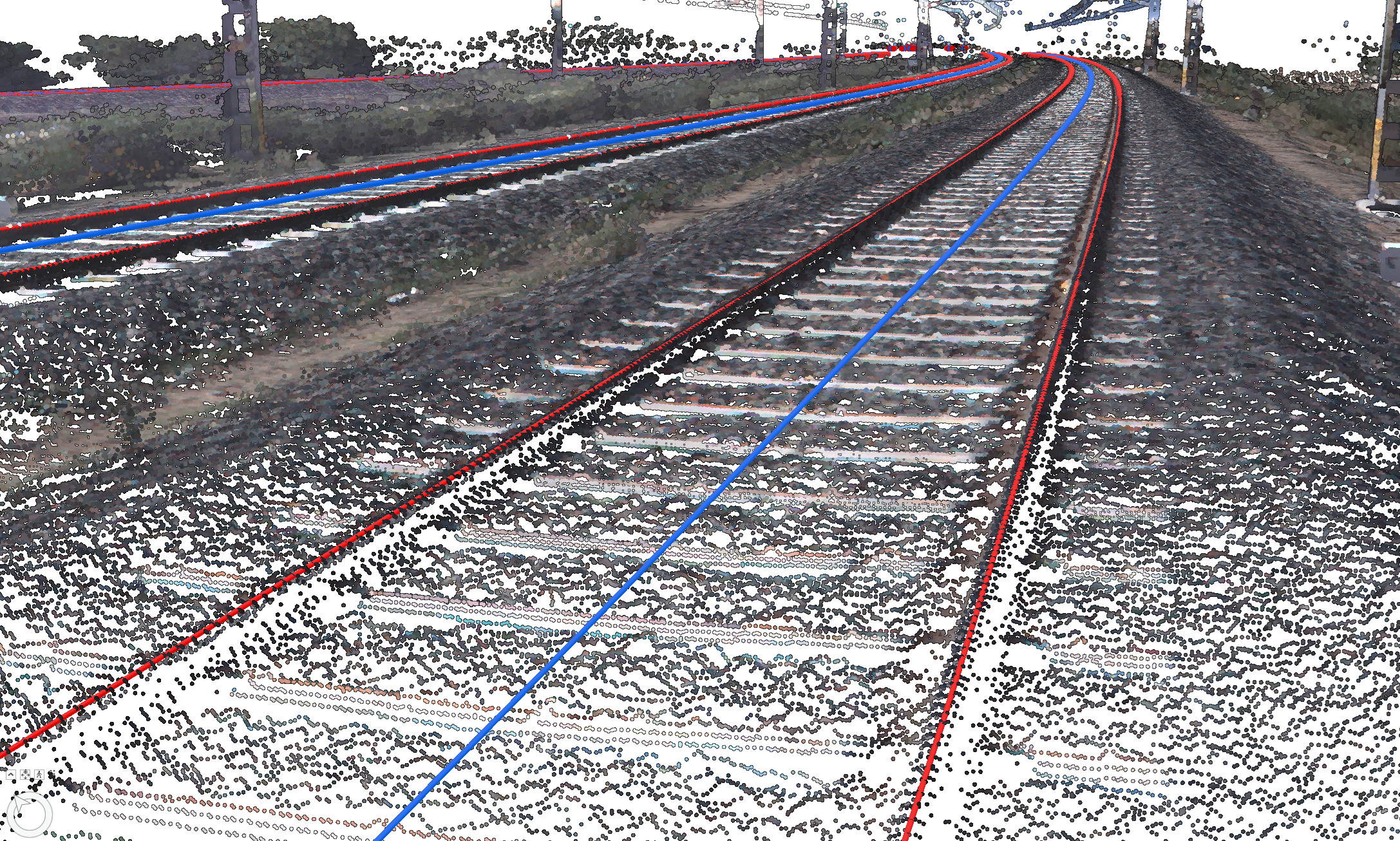}
	\caption
	{
		Example of extracting the railhead top-center polylines (red) for each rail and the resultant track center polyline (blue).  Data courtesy of Esri India.
	}
	\label{fig:example_extract_rail_tracks}
\end{figure}

The example was calculated on the same computer as described in the \secref{sec:QualityEvaluation}. 
Rail reconstruction from point cloud took \SI{1}{\minute} \SI{12}{\second}. Rail lines and centerlines were correctly extracted; see Figure~\ref{fig:example_extract_rail_tracks}.

\section{Conclusion and Future Work\label{sec:ConclusionAndFutureWork}}

\subsection{Conclusion}

This paper presents a novel framework for extracting rail tracks and centerlines from classified mobile lidar point clouds using a raster-based deep learning approach. The method combines rasterization, the recurrent FRPDF neural network, and geometric postprocessing to generate accurate rail polylines and track centerlines. 
The proposed method provides several practical benefits.
\begin{itemize}[label=$\bullet$]
	\item The workflow is highly automated. Once rail points are classified, the algorithm automatically extracts rail geometry and centerlines without manual editing, enabling efficient processing of large railway networks.
	
	\item The approach produces accurate and robust rail models. The FRPDF module operates in a recurrent refinement loop that progressively suppresses noise and restores rail continuity. By both maintaining full spatial resolution and training on diverse synthetic rail scenarios, this approach effectively bridges gaps and preserves the geometric detail required for vectorization.
	
	\item The method does not rely on proprietary training datasets. The neural network is trained entirely on procedurally generated synthetic rail scenarios, avoiding limitations related to data availability and copyright while also eliminating the cost of manually labeling large training datasets.
	
	\item The combination of using a raster-based approach and training on high-noise synthetic data improves the robustness of the algorithm to varying point cloud characteristics. Mobile lidar datasets often exhibit large variations in point density and noise levels due to sensor distance, occlusion, and scanning angle. The algorithm is able to handle these variations across diverse datasets.
\end{itemize}

Experimental results demonstrate that the proposed framework can reliably reconstruct rail geometries and centerlines from noisy lidar point clouds. By automating rail extraction and reducing manual intervention, the method improves the efficiency of railway mapping workflows and supports safer and more effective railway infrastructure management.

\subsection{Future Work}
Although the proposed approach demonstrates promising results, several areas remain for further improvement.
Future work will focus on more complex railway configurations, such as turnouts, rail crossings, and rail yards, where diverging or merging tracks and additional rail components (e.g., switch rails and guard rails) introduce geometric complexity and may interfere with point classification and feature extraction. Extending the synthetic data generation and model training to explicitly include these scenarios could improve the network's ability to recognize and separate branching tracks. Additional postprocessing or specialized submodels may be needed to accurately separate individual rails in switch areas.
In addition, the fidelity of the synthetic training data can be further improved. Incorporating more realistic noise patterns, occlusions, and structural elements into the simulation process may enhance the model's generalizability to challenging real-world conditions.

\subsection{Summary}
This paper presents a raster-based deep learning framework for extracting rail tracks and centerlines from mobile lidar point clouds using synthetic training data. The approach enables efficient and automated reconstruction of rail geometry while avoiding reliance on proprietary datasets as well as the cost of manual labeling.

Beyond railway applications, the combination of a carefully designed neural network and synthetic training data suggests a practical pathway for applying deep learning to other 3D feature extraction and reconstruction tasks. Such approaches could enable more reliable automation of infrastructure mapping workflows, reducing manual processing and improving the efficiency and safety of infrastructure management systems.

\section*{Acknowledgments}

The authors thank Lois Stuart for copyediting the manuscript.

\bibliographystyle{IEEEtran}
\bibliography{ExtractRailsFromPointClouds}

\newcounter{CurrentSectionValue}
\setcounter{CurrentSectionValue}{\value{section}}
\setcounter{section}{0}

\makeatletter
\renewcommand{\theHsection}{labelappendix.\arabic{section}}
\makeatother

\iftrue
\renewcommand{\thesection}{Appendix:}
\else
\renewcommand{\thesection}{Appendix \Roman{section}:}
\fi

\section{Integral of the Squared Distance from a Line Segment to a Point\label{sec:IntegralSegmentToPoint}}

Let \(\overline{\myvector{p}\,\myvector{q}}\) be a line segment of length
$
L \;=\; \|\myvector{q} - \myvector{p}\|.
$
Given a point \(\myvector{r}\), we seek the integral of the squared distance from every point on the segment to \(\myvector{r}\). Without loss of generality, translate \(\myvector{r}\) to the origin (i.e., replace \(\myvector{p}\) and \(\myvector{q}\) with \(\myvector{p} - \myvector{r}\) and \(\myvector{q} - \myvector{r}\)), and assume henceforth that \(\myvector{r} = \myvector{0}\).

Parametrize \(\overline{\myvector{p}\,\myvector{q}}\) by \(t \in [-1,1]\):
\begin{equation}
	\begin{split}
		\myvector{s}\left(t\right) 
		\;=\; 
		\frac{1}{2} \left( \myvector{p} \left(1 + t\right) + \myvector{q} \left(1 - t\right) \right).
	\end{split}
\end{equation}

The integral of \(\|\myvector{s}\left(t\right)\|^2\) over the segment is
\begin{equation}
	\begin{split}
		I 
		&\;=\; 
		\frac{L}{2} \int_{-1}^{1} \|\myvector{s}\left(t\right)\|^2 \, dt \\
		&=\; 
		\frac{L}{2} \int_{-1}^{1} \left\| \frac{ \myvector{p} \left(1 + t\right) + \myvector{q} \left(1 - t\right) }{2} \right\|^2 dt \\
		&=\; 
		\frac{L}{8} \int_{-1}^{1} \|\myvector{p} \left(1 + t\right) + \myvector{q} \left(1 - t\right)\|^2 \, dt \\
		&=\; 
		\frac{L}{8} \int_{-1}^{1} \left( \|\myvector{p}\|^2 \left(1 + t\right)^2 + 2 \left(\myvector{p} \cdot \myvector{q}\right) \left(1 - t^2\right) + \|\myvector{q}\|^2 \left(1 - t\right)^2 \right) dt \\
		&=\; 
		\frac{L}{8} \cdot \frac{8}{3} \left( \|\myvector{p}\|^2 + \myvector{p} \cdot \myvector{q} + \|\myvector{q}\|^2 \right) \\
		&=\; 
		\frac{L}{3} \left( \|\myvector{p}\|^2 + \myvector{p} \cdot \myvector{q} + \|\myvector{q}\|^2 \right).
	\end{split}
\end{equation}

Therefore, the integral of the squared distance from the segment to the \(\myvector{r}\) is 
\begin{equation}
	\boxed{
		I 
		\;=\; 
		\frac{L}{3} \left( \|\myvector{p}\|^2 + \myvector{p} \cdot \myvector{q} + \|\myvector{q}\|^2 \right).
	}
\end{equation}

\end{document}